# Potential Field Guided Actor-Critic Reinforcement Learning

Weiya Ren[1,2]

[1] Artificial Intelligence Research Center of National Innovation Institute of Defense Technology
[2] Tianjin Artificial Intelligence Innovation Center, Tianjin, P.R China.
E-mail: weiyren.phd@gmail.com

**Abstract** In this paper, we consider the problem of actor-critic reinforcement learning. Firstly, we extend the actor-critic architecture to actor-critic-N architecture by introducing more critics beyond rewards. Secondly, we combine the reward-based critic with a potential-field-based critic to formulate the proposed potential field guided actor-critic reinforcement learning approach (actor-critic-2). This can be seen as a combination of the model-based gradients and the model-free gradients in policy improvement. State with large potential field often contains a strong prior information, such as pointing to the target at a long distance or avoiding collision by the side of an obstacle. In this situation, we should trust potential-field-based critic more as policy evaluation to accelerate policy improvement, where action policy tends to be guided. For example, in practical application, learning to avoid obstacles should be guided rather than learned by trial and error. State with small potential filed is often lack of information, for example, at the local minimum point or around the moving target. At this time, we should trust reward-based critic as policy evaluation more to evaluate the long-term return. In this case, action policy tends to explore. In addition, potential field evaluation can be combined with planning to estimate a better state value function. In this way, reward design can focus more on the final stage of reward, rather than reward shaping or phased reward. Furthermore, potential field evaluation can make up for the lack of communication in multi-agent cooperation problem, i.e., multi-agent each has a reward-based critic and a relative unified potential-field-based critic with prior information. Thirdly, simplified experiments on predator-prey game demonstrate the effectiveness of the proposed approach.

## I.  INTRODUCTION

The generalized policy iteration[3], which iterates between policy evaluation and policy improvement, has been adopted by the majority of model-free reinforcement learning algorithms so far. Policy evaluation methods estimate the action-value function, while policy improvement methods update the policy with respect to the action-value function. Based on the generalized policy iteration and the policy gradient theorem, the actor-critic(AC) becomes a widely used architecture[3][4][5]. The deterministic actor-critic algorithms (DPG)[1] further considers deterministic policy gradient algorithms for reinforcement learning with continuous actions. Compared to stochastic policies DPG lowers the variance when estimating the policy gradient. Deep DPG (DDPG)[2] further combines deep neural networks with DPG to improve the modeling capacity. However, both the model-free AC or DDPG algorithms suffer from high sample complexity and low sample efficiency problem.

By using learned model to do imagination rollouts to accelerate the learning or to get better estimates of action-value functions, model-based reinforcement learning methods[6][7][8] allow for more efficient computations and faster convergence than model-free methods. However, the drawback of model-based methods is the model bias. In effect, it is interesting to combine the model-based gradients and the model-free gradients in an efficiently way.

In this paper, we extend the actor-critic architecture to actor-critic-N architecture. As seen in Fig. 1, we define critic1 by reward using deep function approximators. Critic2, Critic3… can be defined by some model-based policy evaluation method. For simplicity, we define Critic2 by potential filed and omit others in this paper.

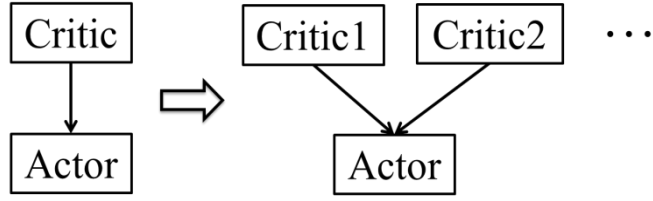

Fig. 1 From Actor-Critic framework to Actor-Critic-N framework. Actor-Critic-2 is illustrated in this paper. Critic1 is based on the reward. Critic2 is based on the potential field.

Consider the example of navigation, the potential value is always large when the robot is far away from the target or very close to the obstacle, and the potential value is always small when the attractive potential value and repulsive potential value are almost equal or the robot is close to the target. We assume that large potential value contains strong prior information. In general, we should trust critic2 more to accelerate learning when potential value is large. In this case, action policy tends to be guided. We should trust critic1 more to reduce the model bias when potential value is small. In this case, action policy tends to explore.

The rest of the paper is organized as follows: Section 2 propose the potential field action-value function. In Section 3 propose the potential field guided actor critic reinforcement learning approach. Experimental results are presented in Section 4. Finally, conclusions are drawn in Section 5.

## II. POTENTIAL FIELD ACTION-VALUE FUNCTION

In classic (artificial) potential field method[11], the most commonly used attractive potential function is:

$$U_{att}(s) = \frac{1}{2}\xi d^2(s, s_{goal}). \tag{1}$$

where, $\xi$ is a positive scaling factor, $d(s, s_{goal})$ is the distance between the point $s$ and the goal $s_{goal}$.

The repulsive potential function is given by

$$U_{att}(s) = \begin{cases} \frac{1}{2}\eta(\frac{1}{d(s, s_{obs})} - \frac{1}{d_0})^2, & if\ d(s, s_{obs}) \leq d_0 \\ 0. & if\ d(s, s_{obs}) > d_0 \end{cases} \tag{2}$$

The overall potential function is given by

$$U(s) = U_{att}(s) + U_{att}(s). \tag{3}$$

In order to define the potential-field-based critic, we first define the **potential field state action function** by

$$q_{PF}(s, a) \triangleq -U(s)[1 - cos(\chi)]. \tag{4}$$

where $\chi$ is the angle between the overall force $f = -\nabla U(s)$ and the actual action $a$ generated by action policy in state $s$, as shown in Fig. 2.

Obviously: If $U(s) \to +\infty$, then $q_{PF}(s,a) \in [0,+\infty)$. If $U(s) \to 0$, then $q_{PF}(s,a) \to 0$.

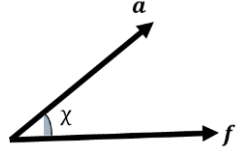

Fig. 2 $\chi$ is the angle between overall force $f$ and actual action $a$.

In this way, $q_{PF}(s,a)$ will play a more important role when the potential value is large and vice versa. In fact, the overall force $f = -\nabla U(s)$ is the simplest single step planning method based on (artificial) potential field. More complex planning methods can be considered, but they are not the focus of this article.

### III. POTENTIAL FIELD GUIDED ACTOR CRITIC REINFORCEMENT LEARNING APPROACH

#### A. Deterministic Policy Gradient Algorithm

We now formally consider a deterministic policy $\mu_\theta: \mathcal{S} \to \mathcal{A}$ with parameter vector $\theta \in R^n$. The reward function is denoted as $r(s,a)$ and the density of initial state distribution is denoted as $p_1(s)$. The density at state $s'$ after transitioning for $t$ time steps from state $s$ is denoted as $p(s \to s', t, \pi)$. The (improper) discounted state distribution is denoted as $\rho^\mu(s',\gamma) = \int_S \sum_{t=1}^{\infty} \gamma^{t-1} p_1(s) p(s \to s', t, \pi) ds$, where $\gamma$ is the discount parameter.

Consider actor-critic-2 and the following objective function

$$J(\mu_\theta) = \beta \int_S \rho^\mu(s,\gamma_1) r(s, \mu_\theta(s)) ds + (1-\beta) \int_S \rho^\mu(s,\gamma_2) q_{PF}(s, \mu_\theta(s)). \tag{5}$$

where $\beta$ is a parameter. When $\beta = 1$, it becomes the classic deterministic policy gradient actor critic algorithm. The first part of the formula is reward-based critic and the second part of the formula is the potential-field-based critic. Notice that these two critics have different discount parameters.

From the deterministic policy gradient theorem[1], we know

$$\nabla_\theta J(\mu_\theta) = \beta E_{s\sim\rho^\mu(s,\gamma_1)}[\nabla_\theta \mu_\theta(s) \nabla_a Q_R^\mu(s,a)|_{a=\mu_\theta(s)}] + (1-\beta) E_{s\sim\rho^\mu(s,\gamma_2)}[\nabla_\theta \mu_\theta(s) \nabla_a Q_{PF}^\mu(s,a)|_{a=\mu_\theta(s)}]. \tag{6}$$

where $Q_R^\mu(s,a) \triangleq E[\sum_{k=1}^{\infty} \gamma_1^{k-1} r(s_k, a_k) | s_1 = s, a_1 = a]$ and $Q_{PF}^\mu(s,a) \triangleq E[\sum_{k=1}^{\infty} \gamma_2^{k-1} q(s_k, a_k) | s_1 = s, a_1 = a]$.

By introducing a function approximator $Q^w(s,a)$, the gradient $\nabla_a Q_R^\mu(s,a)$ can be replaced by $\nabla_a Q^w(s,a)$. Notice that $\gamma_1$ is usually set close to 1, such as $\gamma_1 = 0.99$ due to the long-term reward is important. However, we can simply set $\gamma_2 \to 0$ due to the potential filed can work even without long term expectation. Thus, $\nabla_a Q_{PF}^\mu(s,a)$ can be simply replaced by $\nabla_a q_{PF}(s,a)$. After all, $\nabla_\theta J(\mu_\theta)$ can be approximated by

$$\nabla_\theta J(\mu_\theta) = \beta E_{s\sim\rho^\mu(s,\gamma_1)}[\nabla_\theta \mu_\theta(s) \nabla_a Q^w(s,a)|_{a=\mu_\theta(s)}] + (1-\beta) E_{s\sim\rho^\mu(s,\gamma_2)}[\nabla_\theta \mu_\theta(s) \nabla_a q_{PF}(s,a)|_{a=\mu_\theta(s)}] \tag{7}$$

In the following updating algorithm, Sarsa-update is used to estimate the action-value function[3] in critic1. Action policy is updated by reward-based critic and potential-field-based critic together.

$$\theta_{t+1} \leftarrow \theta_t + \alpha_\theta \nabla_\theta \mu_\theta(s_t)[\beta \nabla_a Q^w(s_t, a_t) + (1-\beta) \nabla q_{PF}(s_t, a_t)]. \tag{8}$$

$$\delta_t \leftarrow R_t + \gamma Q^w(s_{t+1}, a_{t+1}) - Q^w(s_t, a_t). \tag{9}$$

$$w_{t+1} \leftarrow w_t + \alpha_w \delta_t \nabla_w Q^w(s_t, a_t). \tag{10}$$

Notice that $q_{PF}(s, a)$ can also be updated if we give it a parameter. The parameter $\beta$ is set to be 0.5 in the whole paper though it can also be learned dynamically.

### B. Stochastic Policy Gradient

In stochastic policy gradient algorithm, we formally consider a stochastic policy $\pi_\theta$ with parameter vector $\theta \in R^n$. Consider actor-critic-2 and the following objective function

$$J(\pi_\theta) = \beta \int_S \rho^\pi(s, \gamma_1) \int_\mathcal{A} \pi_\theta(s, a) r(s, a) da ds + (1-\beta) \int_S \rho^\pi(s, \gamma_2) \int_\mathcal{A} \pi_\theta(s, a) q_{PF}(s, a)] da ds. \tag{11}$$

By policy gradient theorem[3], we have

$$\nabla_\theta J(\pi_\theta) = \beta \int_S \rho^\pi(s, \gamma_1) \int_\mathcal{A} \nabla_\theta \pi_\theta(s, a) Q_R^\pi(s, a) da ds + (1-\beta) \int_S \rho^\pi(s, \gamma_2) \int_\mathcal{A} \nabla_\theta \pi_\theta(s, a) Q_{PF}^\pi(s, a)] da ds.$$

$$= \beta E_{s \sim \rho^\pi(s, \gamma_1), a \sim \pi_\theta}[\nabla_\theta log \pi_\theta(a|s) Q_R^\pi(s, a)] + (1-\beta) E_{s \sim \rho^\pi(s, \gamma_2), a \sim \pi_\theta}[\nabla_\theta log \pi_\theta(a|s) Q_{PF}^\pi(s, a)]. \tag{12}$$

where $Q_R^\pi(s, a) \triangleq E[\sum_{k=1}^\infty \gamma_1^{k-1} r(s_k, a_k) | s_1 = s, a_1 = a]$ and $Q_{PF}^\pi(s, a) \triangleq E[\sum_{k=1}^\infty \gamma_2^{k-1} q(s_k, a_k) | s_1 = s, a_1 = a]$. The first part of the formula is reward-based critic and the second part of the formula is the potential-field-based critic. Notice that these two critics have different discount parameters.

By introducing a function approximator $V^w(s)$, the $Q_R^\pi(s, a)$ can be replaced by $R + \gamma V^w(s') - V^w(s)$. We also set $\gamma_2 \to 0$. Thus $Q_{PF}^\pi(s, a)$ can be replaced by $q_{PF}(s, a)$. $\nabla_\theta J(\pi_\theta)$ can be approximated by

$$\nabla_\theta J(\pi_\theta) = \beta E_{s \sim \rho^\pi(s, \gamma_1), a \sim \pi_\theta}[\nabla_\theta log \pi_\theta(a|s)(R + \gamma V^w(s') - V^w(s))] + (1-\beta) E_{s \sim \rho^\pi(s, \gamma_2), a \sim \pi_\theta}[\nabla_\theta log \pi_\theta(a|s) q_{PF}(s, a)]. \tag{13}$$

Actor policy and critic1 and can be updated by

$$\theta_{t+1} \leftarrow \theta_t + \alpha_\theta[\beta(R + \gamma V^w(s_{t+1}) - V^w(s_t)) + (1-\beta) q_{PF}(s, a)]] \nabla_\theta log \pi_\theta(a|s). \tag{14}$$

$$\delta_t \leftarrow R_t + \gamma V^w(s_{t+1}) - V^w(s_t). \tag{15}$$

$$w_{t+1} \leftarrow w_t + \alpha_w \delta_t \nabla_w V^w(s_t). \tag{16}$$

As mentioned before, $q_{PF}(s, a)$ can also be updated if we consider parameterized planning approach.

### IV. EXPERIMENT RESULTS

To evaluate the effectiveness of the proposed approach, we introduce a simple predator-prey game based on the OpenAI's multiagent-particle-environments (MPE)[10]. The horizontal and vertical coordinates of the map are limited to [-1, 1] for predators and [-0.8, 0.8] for prey, respectively. Predators and prey have the same maximum speed. In fact, in order to highlight a larger map, the area of predators and prey are relatively small. As shown in Fig. 3. We first consider the case of 1 vs 1 predator-prey game, i.e., one predator pursuits one prey. The environmental rewards are sparse (reward +10 if success) and only depend on the terminal state of each episode. Then, we consider N predators chase one prey in a randomly generated environment. Each time if all predators capture a prey simultaneously, each predator will receive a reward of 10. As long as any predator fails to catch the prey, no reward will be given to any predator. It leads to a difficult learning problem which needs excellent tacit cooperation.

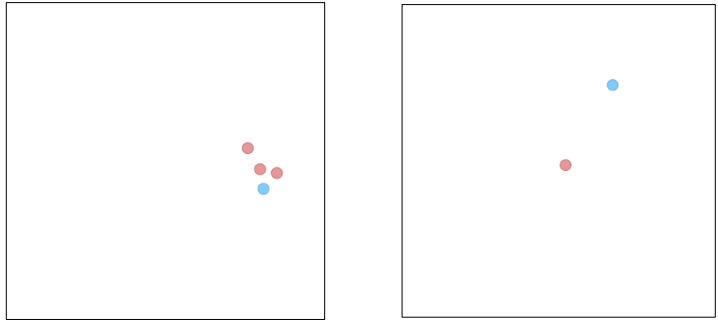

Fig. 3 Illustration of 1vs1 and 3 vs1 predator-prey game.

The target for this experiment is learning to capture the prey independently without knowing each other's policy and action, both in training and testing. For this continuous action control problem, we use the deterministic policy gradient algorithm and denote the proposed potential field guided actor critic reinforcement learning approach as PGDDPG.

For 1 vs 1 predator-prey game, we use a pre-trained DDPG model as the prey's policy. The rate of success capturing (latest 200 episodes) and predator reward curves of PGDDPG versus DDPG are plotted in Fig. 4. To show a smoother learning procedure, the reward value is averaged every 500 episodes. Apparently, the PGDDPG outperforms DDPG in terms of convergence rate.

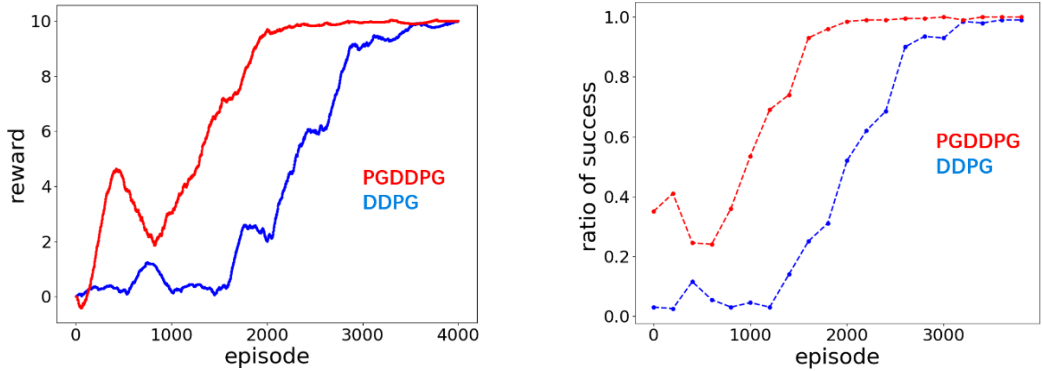

(a) Reward of the predator, which is averaged every 500 episodes.

(b) The ratio of success capturing based on the latest 200 episode.

Fig. 4 1 vs 1 predator-prey game. The red one is the result of PGDDPG vs DDPG, and the blue one is the result of DDPG vs DDPG.

For 3 vs 1 predator-prey game, we use a pre-trained DDPG model as the prey's policy. The rate of success capturing (latest 200 episodes) and predator reward curves of PGDDPG versus DDPG are plotted in Fig. 5. To show a smoother learning procedure, the reward value is averaged every 500 episodes. However, DDPG fails in the 3 vs 1 predator-prey game, while the PGDDPG finally learns how to capture the prey by cooperation. Without communication, the potential-field-based critic can also provide a consistent policy evaluation by a common prior information, which is difficult for reward-based critic.

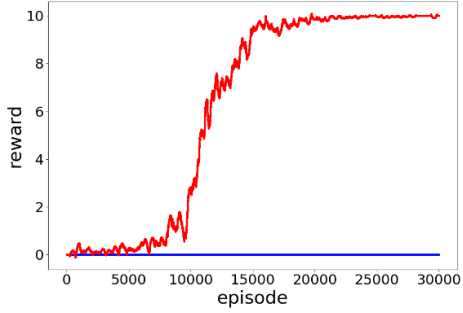
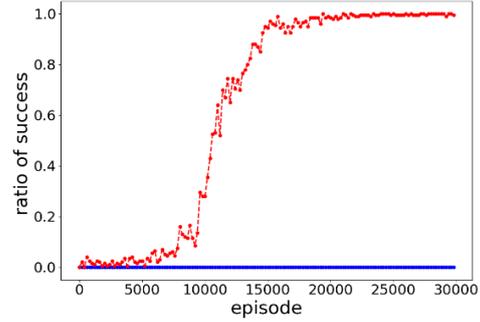

(a)Reward of the predator, which is averaged every 500 episodes.

(b)The ratio of success capturing based on the latest 200 episode.

Fig. 5 3 vs 1 predator-prey game. The red one is the result of PGDDPG vs DDPG, and the blue one is the result of DDPG vs DDPG. The policy of the prey is a pretrained model from PGDDPG vs DDPG.

Instead of using a pre-trained model of prey, we now consider the case of training predators and prey simultaneously. Training simultaneously increases the difficulty of learning, due to it becomes a zero-sum game. As shown in Fig. 6, for PGDDPG, the success rate dropped from 0.2 to 0 at about 1000 episodes (The prey's escape ability exceeds the predator's capture ability). Soon, however, the capture ability caught up with escape ability and took the lead. We can also observe that DDPG fails in the 3 vs 1 predator-prey zero-sum game.

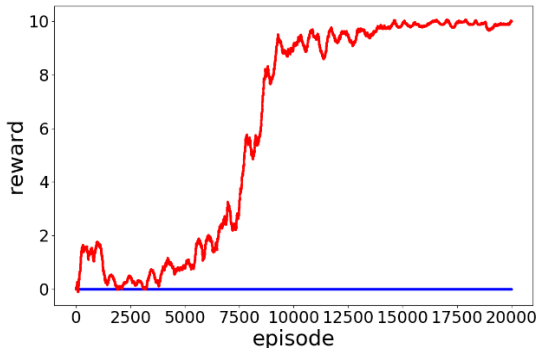
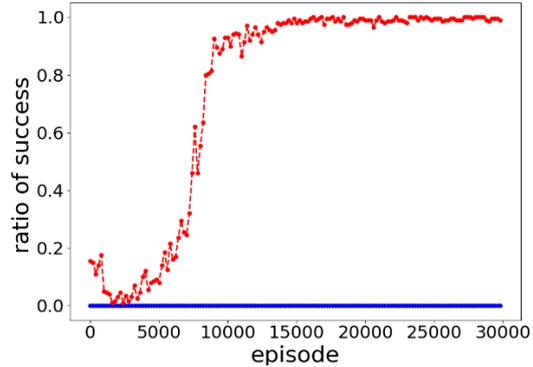

(a)Reward of the predator, which is averaged every 500 episodes.

(b)The ratio of success capturing based on the latest 200 episode.

Fig. 6 3 vs 1 predator-prey game. The red one is the result of PGDDPG vs DDPG, and the blue one is the result of DDPG vs DDPG. The policy of the prey and predators are training simultaneously.

## V. CONCLUSION AND FUTURE WORK

In this paper, we extend the actor-critic architecture to actor-critic-N architecture by introducing model-based policy evaluation methods. We focus on actor-critic-2 and combine the reward-based critic with a potential-field-based critic as the policy evaluation method. We trust the potential-field-based critic more to guide action policy to accelerate learning when the potential field value is relatively large. We trust the reward-based critic more to explore actions and to evaluate the long-term return when the potential field value is relatively small. Reward design can focus more on the final stage of the game, rather than reward shaping or phased reward. Potential field evaluation can make up for the problem of multi-agent cooperation without communication in both training and testing. Furthermore, more planning methods based on the potential field will be studied in the future. Besides, the design of attractive and repulsive potential is also important, e.g., if we have multi goal, we might consider the minimal attractive potential generating from them.

REFERENCES

[1] Silver D, Lever G, Heess N, et al. Deterministic policy gradient algorithms[C]. 2014.
[2] Lillicrap T P, Hunt J J, Pritzel A, et al. Continuous control with deep reinforcement learning[J]. arXiv preprint arXiv:1509.02971, 2015.
[3] Sutton, R. and Barto, A. (1998). Reinforcement Learning: an Introduction. MIT Press.
[4] Peters, J., Vijayakumar, S., and Schaal, S. (2005). Natural actor-critic. In 16th European Conference on Machine Learning, pages 280–291.
[5] Bhatnagar, S., Sutton, R. S., Ghavamzadeh, M., and Lee, M. (2007). Incremental natural actor-critic algorithms. In Neural Information Processing Systems 21.
[6] V. Tangkaratt, S. Mori, T. Zhao, J. Morimoto, and M. Sugiyama. Model-based policy gradients with parameter-based exploration by least-squares conditional density estimation. Neural networks, 57:128–140, 2014.
[7] W. Montgomery and S. Levine. Guided policy search as approximate mirror descent. arXiv preprint arXiv:1607.04614, 2016.
[8] K. Chua, R. Calandra, R. McAllister, and S. Levine. Deep reinforcement learning in a handful of trials using probabilistic dynamics models. In Advances in Neural Information Processing Systems, pages 4754–4765, 2018.
[9] M. Watter, J. Springenberg, J. Boedecker, and M. Riedmiller. Embed to control: A locally linear latent dynamics model for control from raw images. In Advances in neural information processing systems, pages 2746–2754, 2015.
[10] https://github.com/openai/multiagent-particle-envs.
[11] O. Khatib, "Real-time obstacle avoidance for manipulators and mobile robots," Int. J. Rob. Res. 5, 90–98, 1986.